\documentclass{article}
\usepackage{stywhispers,amsmath,epsfig,amsfonts, multirow,hyperref}
\usepackage[
style=ieee,
url=false
]{biblatex}
\title{Implicit Neural Representation for Hyperspectral Video Compression}

\name{Alfredo Scalera, Paul Murray, Jaime Zabalza\thanks{Author email addresses: alfredo.scalera.2019@uni.strath.ac.uk; \{paul.murray, j.zabalza\}@strath.ac.uk}}
\address{University of Strathclyde\\
    Department of Electronic \\
    \& Electrical Engineering\\
    Glasgow, UK}

\begin{document}
\maketitle
\begin{abstract}
With the advent of snapshot cameras, hyperspectral video is becoming more readily available. In recent years, new applications have emerged which have led to increasingly larger datasets. However, hyperspectral video compression remains in the early stages. In this study, we explore the use of implicit neural representation as a candidate solution. We propose a novel extension of an existing RGB video compression model, achieving Bjøntegaard Delta PSNR gains of $+4.99$ dB and Bjøntegaard Delta rate of $-88.88\%$ compared to traditional hyperspectral image compression methods applied frame-by-frame. In addition to reconstruction quality, the effects on downstream task performance are measured in the form of object tracking success. Compared to video compressed with methods based on principal component analysis and JPEG2000 in low data regimes, our proposed method improves tracking area under the curve by up to $23.42\%$ and distance precision by up to $35.56\%$ on examples from the HOT2026 dataset.

\end{abstract}


\section{Introduction}
\label{sec:intro}

Hyperspectral (HS) imagery is commonly used in: quality assessment of food; agriculture; remote sensing; and medical analysis, among others \cite{bhargavaHyperspectralImagingIts2024}. Through technological innovations in camera sensors, HS videos have become more prevalent in recent years. The advent of snapshot hyperspectral sensors has made a number of varied applications emerge, from HS object tracking \cite{xiongMaterialBasedObject2020} to scene segmentation \cite{basterretxeaHSIDriveDatasetResearch2021}. Yet, the field dedicated to HS video compression is still in its infancy and, so far, image compression methods for HS data have been mainly applied frame-by-frame \cite{dasHyperspectralImageVideo2021}.

In this paper a more wholistic approach is taken towards compressing HS video. Consequently, we propose the use of implicit neural representation (INR) for HS video compression \cite{chenHNeRVHybridNeural2023}. INR-based methods leverage the strong data representations enabled by neural networks while avoiding the need for large datasets. 
The main contributions in this paper are: Novel extension of INR-based compression method to HS; Comparison against common PCA-based approach for HS imagery; Quantifying gains from the INR-based method compared to current practices.

\section{Related Work}
\label{sec:related}
Video compression in the domain of RGB video is a mature field in both academic and industrial settings. Most video systems today are built on top of standardised classical hybrid codecs such as AVC \cite{wiegandOverviewH264AVC2003}, HEVC \cite{minallahPerformanceAnalysisH2652015}, and VVC \cite{brossOverviewVersatileVideo2021} which have been developed in the last 25 years.

Over the past decade, with the advent of deep learning, research and development in learned video codecs has emerged. These can broadly be split into two main categories: general and overfit video codecs. General learned video codecs search compressing functions that are learned from end-to-end training on large amounts of data \cite{jiaPracticalRealTimeNeural2025, heEndtoendLearnedVideo2026}.
Overfit video codecs, on the other hand, leverage the overfitting abilities of neural networks thus obtaining an approximation of the functions that generate each pixel or frame of a video. Examples include HNeRV \cite{chenHNeRVHybridNeural2023}, NVRC \cite{kwanNVRCNeuralVideo2024}, and COOL-CHIC \cite{leguayCoolchicVideoLearned2024}.

In recent years, snapshot HS video cameras have emerged due to innovations in mosaic-array sensors \cite{chenHistogramsOrientedMosaic2022}. This has led to the collection and release of various HS video datasets such as HS object tracking (HOT) challenge (HOT2020 \cite{xiongMaterialBasedObject2020}, HOT2026 \cite{hyperspectral-object-tracking-challenge-2026}, IMEC25 \cite{chenHistogramsOrientedMosaic2022}) and automated driving systems (HSI-Drive \cite{basterretxeaHSIDriveDatasetResearch2021}).

HS video compression is often achieved by compressing each datacube frame individually, making use of the spectral correlation between channels \cite{takamuraMultibandVideoCoding2004, pellicer-valeroVideoCompressionSpatiotemporal2025}. The work in \cite{dasHyperspectralImageVideo2021} uses sparse tucker tensor decompositions per-frame, outperforming other HS image compression methods applied in the same manner. Authors in \cite{sippelSyntheticHyperspectralArray2023} propose an HS video coder by building on top of an HS image coder to compensate both spectrally and temporally, achieving higher compression rates than applying only spectral prediction.

To the best of our knowledge, overfit neural compression methods are yet to be applied to the field of HS video compression. In this work we explore their application to the problem and specifically target INR to show the potential of overfit compression.

\section{Methods}
\label{sec:methods}

In the following sections, we detail the methods employed in this study, as well as the experimental setup.

\begin{figure}[htb]

\begin{minipage}[b]{1.0\linewidth}
  \centering
  \centerline{\epsfig{figure="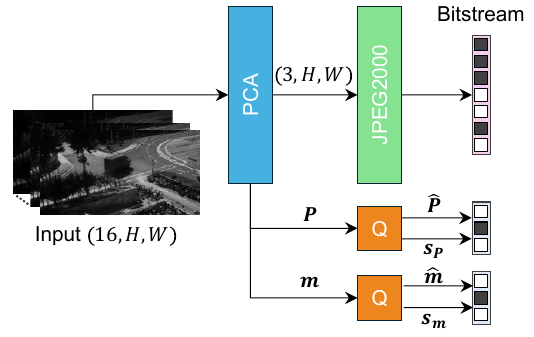",width=8.5cm}}
\end{minipage}

\caption{HS video compression with PCA+JPEG2000. The projection matrix $\boldsymbol{P}$ is quantized to $\hat{\boldsymbol{P}}$ before being written into the bitstream. Q represents min-max quantization and $\boldsymbol{s_P} = (\boldsymbol{P}_\mathrm{min}, \boldsymbol{P}_\mathrm{scale})$ and analogously for $\boldsymbol{m}$.}
\label{fig:pca}

\end{figure}

\subsection{PCA-based Video Compression}
\label{ssec:pcavideo}
PCA is commonly used for spectral dimensionality reduction, and its combination with RGB compression standards has been proposed for HS image and video compression.

Here we explore PCA and JPEG2000 as first proposed in \cite{duHyperspectralImageCompression2007}. The technique is chosen as a strong baseline for overfit HS video compression (since the principal components must be fit to each frame or video). Furthermore, both PCA and JPEG2000 are familiar to the hyperspectral community and readily deployable.

PCA is applied individually to each frame and linearly projects the input datacube $\boldsymbol{D} \in \mathbb{R} ^{C_{\mathrm{in}} \times H \times W}$ from $C_{\mathrm{in}}$ channels to $C_{\mathrm{out}} = 3$ channels to fit JPEG2000. The reduced datacube is then compressed with JPEG2000 to the desired target bits-per-pixel (bpp). The PCA projection matrix $\boldsymbol{P} \in \mathbb{R} ^{C_{\mathrm{in}} \times C_{\mathrm{out}}}$ and mean vector $\boldsymbol{m} \in \mathbb{R}^{C_\mathrm{in}}$ are min-max quantized to improve memory footprint (Eq. \ref{eq:quant}).

\begin{equation}
    q = 
    \Biggl \lfloor{ Q_\mathrm{max} \dfrac
    {\boldsymbol{x} - \boldsymbol{x}_\mathrm{min}}
    {\boldsymbol{x}_\mathrm{scale}}
    }\Biggr \rceil
\label{eq:quant}
\end{equation}

Where $x$ represents the full-precision value, $q$ is the corresponding quantized value, $Q_\mathrm{max}$ the maximum of the quantized range, and $\lfloor \cdot \rceil$ is the rounding operation.
Symmetric signed min-max quantization requires storing the minimum values $\boldsymbol{x}_\mathrm{min} = \min(\boldsymbol{x})$ the scale factors $\boldsymbol{x}_\mathrm{scale} = \max(\boldsymbol{x}) - \min(\boldsymbol{x})$ for dequantization. These are included in the JPEG2000 bitstream as side information $\boldsymbol{s_P} = (\boldsymbol{P}_\mathrm{min}, \boldsymbol{P}_\mathrm{scale}, )$ and $\boldsymbol{s_m} = (\boldsymbol{m}_\mathrm{min}, \boldsymbol{m}_\mathrm{scale}, )$.

When decoding the inverse operations are applied: first the JPEG2000 bitstream is decoded, the projection matrix and mean vector are de-quantized, and finally inverse PCA is applied to the reduced datacube to restore all $C_{\mathrm{in}}$ channels, obtaining the decompressed datacube $\boldsymbol{\tilde{D}}$.

\subsection{INR-based Video Compression}
\label{ssec:inrvideo}
Implicit neural compression can be leveraged in various ways for video compression. Forms include compressed models that learn to map pixel or frame coordinates to pixel values or frame tensors; hybrid schemes where the latent variables and decoders of autoencoder-like networks are compressed \cite{chenHNeRVHybridNeural2023}; and fully end-to-end overfit video codecs \cite{leguayCoolchicVideoLearned2024, kwanNVRCNeuralVideo2024}.

\begin{figure}[htb]

\begin{minipage}[b]{1.0\linewidth}
  \centering
  \centerline{\epsfig{figure="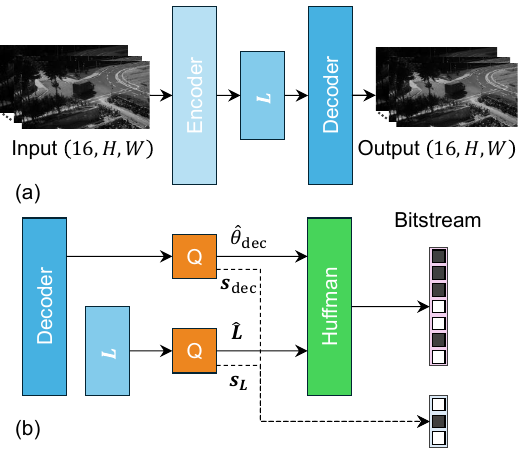",width=8.5cm}}
\end{minipage}

\caption{HS video compression with HNeRV-based method. (a) High-level frame reconstruction for overfitting. (b) Compression - decoder parameters $\theta_{\mathrm{dec}}$ are shared across all video frames while each frame has a unique set of latent variables $\boldsymbol{L}$. Huffman coding losslessly compresses quantized latents $\boldsymbol{\hat{L}}$ and decoder parameters $\hat{\theta}_{\mathrm{dec}}$.}

\label{fig:hnerv}
\end{figure}

In this paper, we adapt HNeRV \cite{chenHNeRVHybridNeural2023} for HS videos. HNeRV is a hybrid RGB approach, treating video compression as a model and latent compression problem (as shown in Fig. \ref{fig:hnerv}).

We apply three extensions to HNeRV: the encoder and decoder networks are adapted to allow multiple channel inputs, $L2$ loss is replaced with bespoke loss $\mathrm{HSFusion}$, and a hyperparameter search was conducted to find a more optimal combination for the HOT2026 dataset. We refer to the hyperspectral version as HS-HNeRV

The $\mathrm{HSFusion}$ loss (Eq. \ref{eq:hsfusion}) combines L2 and cosine-similarity losses. The former targets general reconstruction, while the latter ensures higher spectral fidelity in each pixel.

\begin{equation}
    \mathrm{HSFusion} = \lambda \cdot \mathrm{L2}{\left(\boldsymbol{D}, \boldsymbol{\tilde{D}}\right)} + (1 - \lambda) \cdot \mathrm{Sim}{\left(\boldsymbol{D}, \boldsymbol{\tilde{D}}\right)}
\label{eq:hsfusion}
\end{equation}
Here $\boldsymbol{D}$ and $\boldsymbol{\tilde{D}}$ are the original and reconstructed frame datacubes, respectively. $\mathrm{Sim}$ is cosine-similarity, while $\lambda \in \left[0,1 \right]$ weighs the two losses.

\subsection{Experiments}
\label{ssec:experiments}

\begin{sloppypar}
We measure the effectiveness of INR compared to PCA+JPEG2000 for HS video compression in two ways: (1) reconstruction quality metrics and (2) downstream task performance.    
\end{sloppypar}

PSNR and spectral angle mapper (SAM) are applied to measure the fidelity of the decompressed representations. The first measures general reconstruction, while the latter evaluates the spectral divergence of decoded pixels.

Hyperspectral object tracking (HOT) is chosen as a representative downstream task to measure the impact of compression on computer vision algorithms.
HOT consists of HS videos with moving targets, with the bounding box of a single target provided in the first frame. Algorithms track the object by temporally propagating the bounding box through each frame.
We evaluate CSSTrack \cite{liUnifiedSpatialspectraltemporalNetwork2026}, which represents the current state-of-the-art for HOT tasks. AUC and DP are measured on both raw and decompressed sequences. AUC is the area under the curve of the success plot, which measures the percentage of frames where the intersection-over-union exceeds a particular threshold. DP is distance precision measured as the proportion of frames in which the bounding box center is below a distance error of 20 pixels.

\begin{sloppypar}
Four test video sequences were obtained from the HOT2026 \cite{hyperspectral-object-tracking-challenge-2026} VIS training set \footnote{\texttt{heartsurgery5, high\char`_playground1, high\char`_truck, snowcat}}. The sequences have a native resolution of $512\times272$ pixels and 16 spectral bands and were center cropped to $512\times256$ in all experiments (achieving a 2:1 aspect ratio to accommodate the model).
\end{sloppypar}

HS-HNeRV model sizes and PCA+JPEG2000 target bpp are swept through and PSNR and SAM metrics are computed between the decompressed and original frames for each test sequence. To obtain per-video values, both PSNR and SAM are averaged across all frames.

For performance on HOT tasks, AUC and DP are obtained over three HS-HNeRV model sizes and on PCA+JPEG2000 sequences with matching bpp. We use a CSSTrack checkpoint trained on HOT2020\footnote{\url{https://github.com/hscv/CSSTrack}}.

PCA projection matrix values are quantized to 8-bit unsigned integers after fitting, with minimum and scale factors (side information) embedded as 16-bit floats in the bitstream, as illustrated in Fig. \ref{fig:pca}.

As for HS-HNeRV, the same hyperparameters are set for each video: 64 encoder channels, 16 latent channels, and a learning rate of $1.1\mathrm{e}{-3}$ and trained for 300 epochs with $\lambda = 0.7$.
These were obtained via a random hyperparameter search with 15 trials on four videos from the HOT2026 validation set \footnote{\texttt{athlete1}, \texttt{car8}, \texttt{cat2}, \texttt{pingpong4}}, maximizing peak signal-to-noise ratio (PSNR).
To obtain different bpp values, the model size was varied (3.0, 1.5, 0.75, 0.375 million parameters, respectively).

\section{Results}
\label{sec:results}
\begin{figure}[htb]

\begin{minipage}[b]{1.0\linewidth}
  \centering
  \centerline{\epsfig{figure="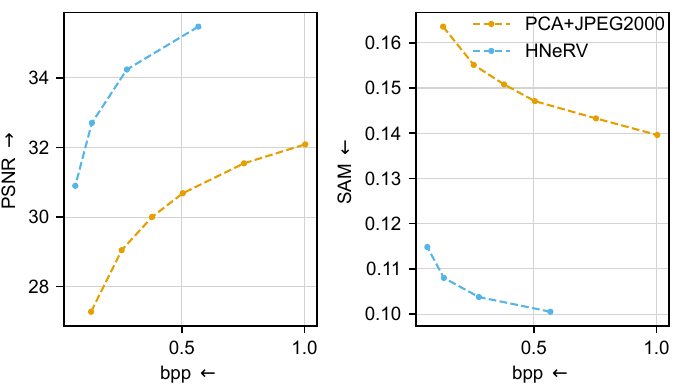",width=8.5cm}}
\end{minipage}

\caption{INR-based compression affords higher fidelity. HS-HNeRV surpasses PCA+JPEG2000 in both spatial (left) and spectral (right) reconstruction at similar compression levels.}
\label{fig:reconstruct}

\end{figure}

Rate-distortion curves are provided in Fig. \ref{fig:reconstruct}. Four HS-HNeRV sizes and six PCA+JPEG2000 bpp targets were evaluated. Values are averaged across all four test sequences. Over equivalent bpp ranges, HS-HNeRV achieves both higher PSNR and lower SAM scores. Specifically, HS-HNeRV achieves a Bjøntegaard Delta (BD)-rate of $-88.88 \%$ and a BD-PSNR of $4.99$ dB with respect to PCA+JPEG2000.

\begin{sloppypar}
In low bitrate regimes, HS-HNeRV also outperforms PCA+JPEG2000 in maintaining tracking performance, as reported in Table \ref{tab:tracking}.
For bpp of approx.\ $0.13$, HS-HNeRV improves over the former by $23.42\%$ and $35.56\%$ in tracking AUC and DP, respectively.
On the other hand, PCA+JPEG2000 surpasses HS-HNeRV at lower compression rates. Interestingly, the DP of CSSTrack improves on sequences compressed with PCA+JPEG2000 at $0.6$ bpp compared to the raw data.
\end{sloppypar}

\begin{table}[htb]
\caption{CSSTrack performance on compressed HOT2026.}
\medskip
\label{tab:tracking}
\begin{center}
\begin{tabular}{c|c|c|c}
\hline
\textbf{Compression Method}&\textbf{bpp}&\textbf{AUC}&\textbf{DP} \\
\hline
Raw Sequence&   --&     $43.59$&    $67.84$ \\
\hline
\multirow{3}{*}{PCA+JPEG2000}&   0.135&  $24.55$&	$42.80$ \\
&   0.275&  $27.87$&	$46.44$ \\
&   0.604&  $37.92$&    $69.92$ \\
\hline
\multirow{3}{*}{HS-HNeRV}&  0.133&  $30.30$&    $58.02$ \\
&  0.276&  $33.37$&    $61.24$ \\
&  0.568&  $35.02$&    $62.48$ \\
\hline
\end{tabular}
\end{center}
\end{table}

\section{Discussion}
\label{sec:discussion}

\begin{figure}[htb]

\begin{minipage}[b]{1.0\linewidth}
  \centering
  \centerline{\epsfig{figure="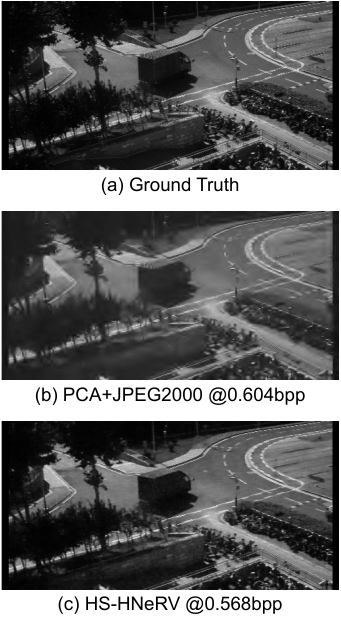",width=7.75cm}}
\end{minipage}

\caption{Qualitative comparison of reconstruction capabilities. Channel 16 of first frame from \texttt{high\char`_truck} sequence. HS-HNeRV achieves a higher spatial and spectral quality.}
\label{fig:quality}

\end{figure}

INR-based methods show clear potential for HS video compression compared to PCA-based overfit approaches. Stronger representation is seen across all bitrates with HS-HNeRV (example illustrated in Fig. \ref{fig:quality}), achieving the same PSNR as PCA+JPEG2000 while requiring only $11.12\%$ of the data (BD-rate of $-88.88\%$). The potential for low-data environments is also demonstrated by the increased HOT performance.
Overall, an INR-based overfit approach removes the need for large amounts of data for a general model while leveraging the function fitting capabilities of deep learning.

\begin{figure}[htb]

\begin{minipage}[b]{1.0\linewidth}
  \centering
  \centerline{\epsfig{figure="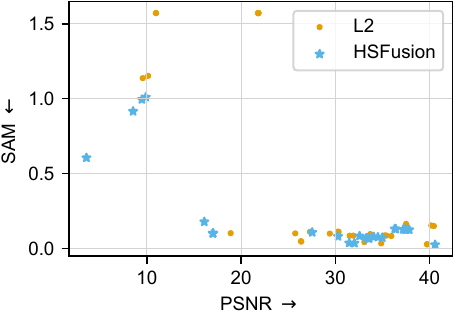",width=8.25cm}}
\end{minipage}

\caption{Generally, for equivalent PSNR values, HSFusion forces better spectral fidelity within each pixel (SAM metric). Each datum represents a hyperparameter search trial.}
\label{fig:losses}

\end{figure}

The proposed $\mathrm{HSFusion}$ loss is also beneficial for algorithms that operate at the pixel-level as it forces spectral consistency compared to the standard $\mathrm{L2}$ loss, as seen in Fig. \ref{fig:losses}.

Higher compression ratios could be further achieved by employing more advanced quantization and encoding schemes. For example, quantization-aware training and compact entropy models. 
The performance of HNeRV and similar methods is highly dependent on the right choice of hyperparameters. General hyperparameters can be found via search strategies on multiple videos, but this defeats the purpose of dataset-free compression when input distributions differ. This can be mitigated by adding hyperparameter optimization strategies, such as Bayesian or gradient-based optimization. 

\section{Conclusion}
\label{sec:conclusion}
In this work, we demonstrate the way in which INR can be leveraged for HS video compression. The methods shown demonstrate strong performance compared to PCA combined with JPEG2000 for video encoding, improving reconstruction quality by $+4.99$ dB.

Future work will include the adoption of bespoke architectures for HS video, the use of hyperparameter optimization for ideal overfitting, and adjusting quantization and encoding procedures to match the latest overfitting RGB codecs.

\section{REFERENCES}
\label{sec:ref}

\begin{sloppypar}
    \printbibliography[heading=none]

@article{bhargavaHyperspectralImagingIts2024,
  title = {Hyperspectral Imaging and Its Applications: {{A}} Review},
  shorttitle = {Hyperspectral Imaging and Its Applications},
  author = {Bhargava, Anuja and Sachdeva, Ashish and Sharma, Kulbhushan and Alsharif, Mohammed H. and Uthansakul, Peerapong and Uthansakul, Monthippa},
  date = {2024-06-30},
  journaltitle = {Heliyon},
  shortjournal = {Heliyon},
  volume = {10},
  number = {12},
  pages = {e33208},
  issn = {2405-8440},
  doi = {10.1016/j.heliyon.2024.e33208},
  url = {https://www.sciencedirect.com/science/article/pii/S2405844024092399},
  urldate = {2026-07-22}
}

@misc{hyperspectral-object-tracking-challenge-2026,
    author = {HotTracking2026},
    title = {Hyperspectral Object Tracking Challenge 2026},
    year = {2026},
    howpublished = {\url{https://kaggle.com/competitions/hyperspectral-object-tracking-challenge-2026}},
    note = {Kaggle}
}

@article{xiongMaterialBasedObject2020,
  title = {Material {{Based Object Tracking}} in {{Hyperspectral Videos}}},
  author = {Xiong, Fengchao and Zhou, Jun and Qian, Yuntao},
  date = {2020},
  journaltitle = {IEEE Transactions on Image Processing},
  volume = {29},
  pages = {3719--3733},
  issn = {1941-0042},
  doi = {10.1109/TIP.2020.2965302},
  url = {https://ieeexplore.ieee.org/abstract/document/8960632},
  urldate = {2026-05-21}
}

@inproceedings{basterretxeaHSIDriveDatasetResearch2021,
  title = {{{HSI-Drive}}: {{A Dataset}} for the {{Research}} of {{Hyperspectral Image Processing Applied}} to {{Autonomous Driving Systems}}},
  shorttitle = {{{HSI-Drive}}},
  booktitle = {2021 {{IEEE Intelligent Vehicles Symposium}} ({{IV}})},
  author = {Basterretxea, K. and Martínez, V. and Echanobe, J. and Gutiérrez–Zaballa, J. and Del Campo, I.},
  date = {2021-07},
  pages = {866--873},
  doi = {10.1109/IV48863.2021.9575298},
  url = {https://ieeexplore.ieee.org/abstract/document/9575298},
  urldate = {2026-07-22},
  eventtitle = {2021 {{IEEE Intelligent Vehicles Symposium}} ({{IV}})}
}

@article{duHyperspectralImageCompression2007,
  title = {Hyperspectral {{Image Compression Using JPEG2000}} and {{Principal Component Analysis}}},
  author = {Du, Qian and Fowler, James E.},
  date = {2007-04},
  journaltitle = {IEEE Geoscience and Remote Sensing Letters},
  volume = {4},
  number = {2},
  pages = {201--205},
  issn = {1558-0571},
  doi = {10.1109/LGRS.2006.888109},
  url = {https://ieeexplore.ieee.org/abstract/document/4156154},
  urldate = {2026-06-16}
}

@online{chenHNeRVHybridNeural2023,
  title = {{{HNeRV}}: {{A Hybrid Neural Representation}} for {{Videos}}},
  shorttitle = {{{HNeRV}}},
  author = {Chen, Hao and Gwilliam, Matt and Lim, Ser-Nam and Shrivastava, Abhinav},
  date = {2023-04-05},
  url = {https://arxiv.org/abs/2304.02633v1},
  urldate = {2026-06-03},
  langid = {english},
  pubstate = {prepublished}
}

@article{wiegandOverviewH264AVC2003,
  title = {Overview of the {{H}}.264/{{AVC}} Video Coding Standard},
  author = {Wiegand, T. and Sullivan, G.J. and Bjontegaard, G. and Luthra, A.},
  date = {2003-07},
  journaltitle = {IEEE Transactions on Circuits and Systems for Video Technology},
  volume = {13},
  number = {7},
  pages = {560--576},
  issn = {1558-2205},
  doi = {10.1109/TCSVT.2003.815165},
  url = {https://ieeexplore.ieee.org/abstract/document/1218189},
  urldate = {2025-10-27}
}

@article{brossOverviewVersatileVideo2021,
  title = {Overview of the {{Versatile Video Coding}} ({{VVC}}) {{Standard}} and Its {{Applications}}},
  author = {Bross, Benjamin and Wang, Ye-Kui and Ye, Yan and Liu, Shan and Chen, Jianle and Sullivan, Gary J. and Ohm, Jens-Rainer},
  date = {2021-10},
  journaltitle = {IEEE Transactions on Circuits and Systems for Video Technology},
  volume = {31},
  number = {10},
  pages = {3736--3764},
  issn = {1558-2205},
  doi = {10.1109/TCSVT.2021.3101953},
  url = {https://ieeexplore.ieee.org/document/9503377},
  urldate = {2025-12-05}
}

@inproceedings{minallahPerformanceAnalysisH2652015,
  title = {Performance {{Analysis}} of {{H}}.265/{{HEVC}} ({{High-Efficiency Video Coding}}) with {{Reference}} to {{Other Codecs}}},
  booktitle = {2015 13th {{International Conference}} on {{Frontiers}} of {{Information Technology}} ({{FIT}})},
  author = {Minallah, N. and Gul, S. and Bokhari, M.M.},
  date = {2015-12},
  pages = {216--221},
  doi = {10.1109/FIT.2015.46},
  url = {https://ieeexplore.ieee.org/document/7421003},
  urldate = {2026-07-22},
  eventtitle = {2015 13th {{International Conference}} on {{Frontiers}} of {{Information Technology}} ({{FIT}})}
}

@article{heEndtoendLearnedVideo2026,
  title = {End-to-End Learned Video Compression: {{A}} Comprehensive Review},
  shorttitle = {End-to-End Learned Video Compression},
  author = {He, Huanjie and Shi, Yunhui and Wang, Jin and Liang, Jiuzhen and Ling, Nam and Yin, Baocai},
  date = {2026-09-14},
  journaltitle = {Neurocomputing},
  shortjournal = {Neurocomputing},
  volume = {694},
  pages = {133839},
  issn = {0925-2312},
  doi = {10.1016/j.neucom.2026.133839},
  url = {https://www.sciencedirect.com/science/article/pii/S0925231226012361},
  urldate = {2026-05-20}
}

@online{jiaPracticalRealTimeNeural2025,
  title = {Towards {{Practical Real-Time Neural Video Compression}}},
  author = {Jia, Zhaoyang and Li, Bin and Li, Jiahao and Xie, Wenxuan and Qi, Linfeng and Li, Houqiang and Lu, Yan},
  date = {2025-02-28},
  url = {https://arxiv.org/abs/2502.20762v2},
  urldate = {2026-06-03},
  langid = {english},
  pubstate = {prepublished}
}

@inproceedings{kwanNVRCNeuralVideo2024,
  title = {{{NVRC}}: {{Neural Video Representation Compression}}},
  shorttitle = {{{NVRC}}},
  booktitle = {Advances in {{Neural Information Processing Systems}}},
  author = {Kwan, Ho Man and Gao, Ge and Zhang, Fan and Gower, Andrew and Bull, David},
  date = {2024},
  volume = {37},
  pages = {132440--132462},
  publisher = {Curran Associates, Inc.},
  doi = {10.52202/079017-4210},
  url = {https://proceedings.neurips.cc/paper_files/paper/2024/hash/eed57814c16645298db3164829e2e45c-Abstract-Conference.html},
  urldate = {2026-07-14}
}

@inproceedings{leguayCoolchicVideoLearned2024,
  title = {Cool-Chic Video: {{Learned}} Video Coding with 800 Parameters},
  shorttitle = {Cool-Chic Video},
  booktitle = {2024 {{Data Compression Conference}} ({{DCC}})},
  author = {Leguay, Thomas and Ladune, Théo and Philippe, Pierrick and Déforges, Olivier},
  date = {2024-03},
  pages = {23--32},
  issn = {2375-0359},
  doi = {10.1109/DCC58796.2024.00010},
  url = {https://ieeexplore.ieee.org/abstract/document/10533789},
  urldate = {2026-06-26},
  eventtitle = {2024 {{Data Compression Conference}} ({{DCC}})}
}

@article{chenHistogramsOrientedMosaic2022,
  title = {Histograms of Oriented Mosaic Gradients for Snapshot Spectral Image Description},
  author = {Chen, Lulu and Zhao, Yongqiang and Chan, Jonathan Cheung-Wai and Kong, Seong G.},
  date = {2022-01-01},
  journaltitle = {ISPRS Journal of Photogrammetry and Remote Sensing},
  shortjournal = {ISPRS Journal of Photogrammetry and Remote Sensing},
  volume = {183},
  pages = {79--93},
  issn = {0924-2716},
  doi = {10.1016/j.isprsjprs.2021.10.018},
  url = {https://www.sciencedirect.com/science/article/pii/S0924271621002860},
  urldate = {2026-07-22}
}

@inproceedings{takamuraMultibandVideoCoding2004,
  title = {Multiband Video Coding Using {{H}}.264/{{AVC}}, {{MPEG-4}} Studio Profile and {{JPEG}} 2000},
  booktitle = {Data {{Compression Conference}}, 2004. {{Proceedings}}. {{DCC}} 2004},
  author = {Takamura, S. and Yashima, Y.},
  date = {2004-03},
  pages = {567-},
  issn = {1068-0314},
  doi = {10.1109/DCC.2004.1281543},
  url = {https://ieeexplore.ieee.org/abstract/document/1281543},
  urldate = {2026-04-08},
  eventtitle = {Data {{Compression Conference}}, 2004. {{DCC}} 2004}
}

@online{pellicer-valeroVideoCompressionSpatiotemporal2025,
  title = {Video {{Compression}} for {{Spatiotemporal Earth System Data}}},
  author = {Pellicer-Valero, Oscar J. and Aybar, Cesar and Valls, Gustau Camps},
  date = {2025-06-24},
  eprint = {2506.19656},
  eprinttype = {arXiv},
  eprintclass = {cs},
  doi = {10.48550/arXiv.2506.19656},
  url = {http://arxiv.org/abs/2506.19656},
  urldate = {2026-04-10},
  pubstate = {prepublished}
}

@article{dasHyperspectralImageVideo2021,
  title = {Hyperspectral Image, Video Compression Using Sparse Tucker Tensor Decomposition},
  author = {Das, Samiran},
  date = {2021-03-01},
  journaltitle = {IET Image Processing},
  volume = {15},
  number = {4},
  pages = {964--973},
  publisher = {John Wiley \& Sons, Ltd},
  issn = {1751-9667},
  doi = {10.1049/ipr2.12077},
  url = {https://ietresearch.onlinelibrary.wiley.com/doi/10.1049/ipr2.12077},
  urldate = {2026-04-10},
  langid = {english}
}

@article{sippelSyntheticHyperspectralArray2023,
  title = {Synthetic Hyperspectral Array Video Database with~Applications to Cross-Spectral Reconstruction and~Hyperspectral Video Coding},
  author = {Sippel, Frank and Seiler, Jürgen and Kaup, André},
  date = {2023-03-01},
  journaltitle = {JOSA A},
  shortjournal = {J. Opt. Soc. Am. A, JOSAA},
  volume = {40},
  number = {3},
  pages = {479--491},
  publisher = {Optica Publishing Group},
  issn = {1520-8532},
  doi = {10.1364/JOSAA.479552},
  url = {https://opg.optica.org/josaa/abstract.cfm?uri=josaa-40-3-479},
  urldate = {2026-04-08},
  langid = {english}
}

@article{liUnifiedSpatialspectraltemporalNetwork2026,
  title = {A Unified Spatial-Spectral-Temporal Network for Hyperspectral Object Tracking},
  author = {Li, Zhuanfeng and Wang, Jing and Zhang, Jue and Zhao, Dong and Fu, Guanyiman and Wang, Jiangtao and Lu, Jianfeng},
  date = {2026-06-01},
  journaltitle = {Pattern Recognition},
  shortjournal = {Pattern Recognition},
  volume = {174},
  pages = {113005},
  issn = {0031-3203},
  doi = {10.1016/j.patcog.2025.113005},
  url = {https://www.sciencedirect.com/science/article/pii/S0031320325016681},
  urldate = {2026-07-07}
}
\end{sloppypar}

\end{document}